# Image-based monitoring of bolt loosening through deep-learning-based integrated detection and tracking

Xiao Pan[a], T.Y. Yang[b]

[a] PhD candidate, Department of Civil Engineering, University of British Columbia, Vancouver, Canada (p.xiao1994@outlook.com)

[b] Professor, Department of Civil Engineering, University of British Columbia, Vancouver, Canada (yang@civil.ubc.ca)

**ABSTRACT:** Structural bolts are critical components used in different structural elements, such as beam-column connections and friction damping devices. The clamping force in structural bolts is highly influenced by the bolt rotation. Much of the existing vision-based research about bolt rotation estimation relies on traditional computer vision algorithms such as Hough Transform to assess static images of bolts. This requires careful image preprocessing, and it may not perform well in the situation of complicated bolt assemblies, or in the presence of surrounding objects and background noise, thus hindering their real-world applications. In this study, an integrated real-time detect-track method, namely RTDT-Bolt, is proposed to monitor the bolt rotation angle. First, a real-time convolutional-neural-networks-based object detector, named YOLOv3-tiny, is established and trained to localize structural bolts. Then, the target-free object tracking algorithm based on optical flow is implemented, to continuously monitor and quantify the rotation of structural bolts. In order to enhance the tracking performance against background noise and potential illumination changes during tracking, the YOLOv3-tiny is integrated with the optical flow tracking algorithm to re-detect the bolts when the tracking gets lost. Extensive parameter studies were conducted to identify optimal tracking performance and examine the potential limitations. The results indicate the RTDT-Bolt method can greatly enhance the tracking performance of bolt rotation, which can achieve over 90% accuracy using the recommended range for the parameters.

## 1 INTRODUCTION

Structural bolts are critical parts to connect structural elements in place. Structural components such as beams-column joints and column-base connections can experience complete failure if the bolts get loosened to a certain level, which may result in a catastrophic system-level collapse. Besides, some of the innovative energy dissipation devices such as friction dampers heavily rely on the bolts to generate the desired friction force. The seismic energy absorption potential of such damping devices will deteriorate with the loosening of bolts, and consequently, affect the global performance of the building. Therefore, robust monitoring methods should be developed to detect damages in bolted components, and if the damage has been found, repair or replacement actions should be applied to maintain the structural integrity, prior to extreme natural hazards.

Earlier, traditional structural health monitoring (SHM) methods (Salawu, 1997; Wang, Song, Liu, et al., 2013; Qarib, & Adeli, 2014) were developed to replace time-consuming manual inspections. In general, SHM methods using contact type sensors identify damage based on the structural modal properties (i.e., stiffness and damping), which are related to natural frequencies and mode shapes. The contact-sensor-based SHM methods to identify bolt loosening have also been developed (Yang, & Chang, 2006; Wang et al., 2013; Sevillano, Sun, & Perera, 2016). However, these contract sensor-based methods have several limitations. Contact sensors are unreliable when subjected to the change of environmental conditions, such as temperature and humidity, which could lead to false detection (Xia, Chen, Weng, et al., 2012; Li, Deng, & Xie, 2015). These methods require dedicated experts to set up the sensors, high-precision instrumentation, and a software package to account for environmental variation effects (Huynh, & Kim, 2017; Huynh, & Kim, 2018). Moreover, in case of bolt loosening detection, these methods could recognize damage in the bolted assemblies, but could not precisely localize the loosened bolts (Ramana, Choi, & Cha, 2019). Such methods are labor-intensive, expensive, and may be impractical in real-world applications to assess the bolt loosening in a device with a large number of bolts of different types, which requires many sensors to be set up differently.

In recent years, vision-based SHM has evolved as an reliable and efficient way for structural damage detection in various civil engineering applications. In comparison to the contact-type sensors, vision-based methods use non-contact detection and have low cost in sensors, and their installation and operation. For example, a simple commercial-grade camera can capture multiple bolts at the same time, and the data acquired can be efficiently processed and analyzed by modern computers. In addition, due to the nature of cameras, the damage features in the image are immune to changes of temperature or humidity. In recent years, convolutional neural networks (CNNs), which falls into a category of deep neural networks (or deep learning), have been shown to prominently outperform the traditional image processing techniques (IPTs). Notably, Advanced CNNs models such as AlexNet (Krizhevsky, Sutskever, & Hinton, 2012), VGG (Simonyan, & Zisserman, 2014), GoogleNet (Szegedy et al., 2015), ResNet (He, Zhang, Ren, & Sun, 2016), DenseNet (Huang, Liu, Van Der Maaten, & Weinberger, 2017), DarkNet (Redmon, & Farhadi, 2018) have been developed and optimized in speed and accuracy in image classification tasks. CNNs have been effectively implemented in health monitoring of various structural components (Cha, Choi, & Büyüköztürk, 2017; Xu, Gui, & Han, 2020; Azimi, & Pekcan, 2020; Miao, Ji, Okazaki, & Takahashi, 2021; Gao, Zhai, & Mosalam, 2021; Sajedi, & Liang, 2021). Moreover, other noticeable developments of deep learning in various civil engineering applications have been attempted, such as damage localization of frame-type structures using decision tree ensemble method (Mariniello et al., 2021), damage detection of experimental wooden bridge using unsupervised learning method based on kernel null space and probabilistic threshold estimation (Sarmadi, & Yuen, 2021), damage detection of civil structures using synchro-squeezed wavelet transform and fractality dimension method (Amezquita-Sanchez, & Adeli, 2015; Perez-Ramirez et al., 2016; Li, Park, & Adeli, 2017), development of sustainable strain-sensing model using radial basis function neural network and genetic algorithm (Oh et al., 2017), structural response prediction of high-rise buildings using recurrent neural network with Bayesian regularization (Perez-Ramirez et al., 2019), estimation of concrete properties using deep restricted Boltzmann machine (Rafiei, Khushefati, Demirboga, & Adeli, 2017a), structural health condition assessment of high rise buildings through neural dynamics classification (Rafiei, & Adeli, 2017b), development of seismicity indicator-based earthquake early warning system (Rafiei, & Adeli, 2017c), structural condition assessment using synchrosqueezed wavelet transform and unsupervised deep Boltzmann machine (Rafiei, & Adeli, 2018), image-based regional post-disaster damage assessment using time-frequency distribution diagrams of ground motions (Lu, 2021) and long short-term memory neural networks (Xu, Lu, 2021).

Further, CNNs have been dominant in the field of image-based object detection, which incorporates the classification and localization of an object. Region-based CNN methods, including RCNN (Girshick, Donahue, Darrell, & Malik, 2014), Fast-RCNN (Girshick, 2015) and Faster-RCNN (Ren, He, Girshick, & Sun, 2015) and Mask RCNN (He, Gkioxari, Dollár, & Girshick, 2017) have been developed for image-based object detection and deployed in civil engineering applications. On the other hand, the regression-based detector, YOLO (version 1) algorithm was first introduced by Redmon, Divvala, Girshick, & Farhadi, (2016) for real-time object detection, which is much faster than the RCNN method. Further, YOLOv2 (Redmon, & Farhadi, 2017), YOLOv3 (Redmon, & Farhadi, 2018), YOLOv4 (Bochkovskiy, Wang, & Liao, 2020), and YOLOv5 (Jocher, 2020), was developed to enhance the localization precision particularly for small objects. Many of these algorithms have been successfully applied in structural damage detection. For example, Cha, Choi, Suh, Mahmoudkhani, and Büyüköztürk (2018) trained the Faster-RCNN to detect various structural damage types for steel and concrete structures. Liang (2019) applied Faster-RCNN for RC bridge column localization. Cheng, Behzadan, and Noshadravan (2021) employed Mask R-CNN to localize the buildings for post-hurricane damage assessment. In addition to region-based CNN methods, other methods have been attempted for damage localization. Pan and Yang (2020) implemented a regression-based detection algorithm, YOLOv2, to detect exposed steel reinforcements and estimate the repair cost for reinforced concrete structures. Wu (2021) proposed a rail boundary guidance network which is integrated with YOLOv3 to detect the surface defects of railways. Jiang, Han, Du and Ni (2021) proposed a diagnostic framework using deep auto-encoder and manifold learning for damage localization and quantification of a benchmark steel frame.

To date, a limited number of vision-based studies for bolt loosening detection were conducted, such as the employment of Hough Transform (HT) algorithm (Hart, & Duda, 1972) to detect bolt loosening (Park, Kim, & Kim, 2015a; Park, Huynh, Choi, & Kim, 2015b), the integration of HT and support vector machine for bolt loosening quantification (Cha et al., 2016; Ramana et al., 2017; Ramana et al., 2019), and image registration approach for bolt loosening angle estimation (Kong & Li, 2018). Recently, Huynh et al. (2019) employed HT and R-CNN algorithms to estimate the bolt rotation angle, which was validated on a full-scale bridge connection using an unmanned aerial vehicle. Ta and Kim (2020) further applied similar algorithms to detect bolt loosening and corrosion. Although the results have been promising, their methods bear three main limitations. To wit: a) the methodology relied on HT algorithm to detect lines and circles, which as aforementioned, may not perform well when the bolt assembly becomes more complicated, with the existence of washers, or in the situation of light reflection on the bolts, shades of the surrounding objects on the bolts, or the presence of background noise in reality; b) the estimation of rotation angle is conducted on two static images, using geometric transformation analysis of edge lines of bolts. This only works effectively when the rotation angle between these two images is less than 60 degrees, due to the geometric nature of hexagon-shaped bolt nuts examined; c) despite the RCNN method provides a reasonable accuracy with proper training, the speed of RCNN is too slow for real-time applications (Redmon, & Farhadi, 2018). To be more specific, the rotational speed of bolts under severe earthquake shaking can be relatively high (e.g., up to 90 degrees per second). In order to ensure the tracking method can track bolt rotation, it requires the rotation of bolts to be less than 60 degrees between two adjacent video frames. This means the minimum processing speed for real-time detection and tracking of bolt rotation is 1.5 FPS (= 90 degrees per second / 60 degrees per frame), which cannot be achieved by RCNN method.

To address these issues, in this paper, an integrated real-time detect-track (RTDT) method, named RTDT-bolt, for bolt rotation is proposed. To the best of the authors' knowledge, the proposed method is the first-of-its-kind to detect and track the bolt rotation interactively, using vision-based techniques. The implementation has been briefly summarized herein, and the details will be explained in Section 2.3. First, the object detection algorithm, YOLOv3-tiny, which was built upon the original architecture of YOLOv3 (Redmon, & Farhadi, 2018), was trained for accurate real-time bolt localization in an image, under various lighting conditions. YOLOv3-tiny has a reduced depth of the convolutional layers compared to the original YOLOv3, thus greatly improving the detection speed, while still maintaining competitive accuracy. Second, feature points (FPs) are generated using the Shi-Tomasi corner detection algorithm (Shi, &Tomasi, 1994) within the regions of interest (ROIs) detected by YOLOv3-tiny, and tracked through Kanade-Lucas-Tomasi (KLT) optical flow tracking algorithm (Tomasi, & Kanade, 1991). The geometric transformation analysis based on M-estimator Sample Consensus (MSAC) algorithm (Torr, & Zisserman, 2000) is then developed to estimate the frame-to-frame rotation. Third, given that optical flow algorithm tends to fail in the presence of sudden change in pixel values, likely due to illumination changes and accumulated errors from background noise during tracking (Nixon, & Aguado, 2019), to address this issue, the tracking algorithm is integrated with the YOLOv3-tiny algorithm to re-detect the target when the tracking gets lost. The tracking continues with the new FPs generated every time when the new detection is imposed. The proposed method can allow the users to continuously track the bolt rotation in real time. To demonstrate the effectiveness and examine the potential limitations of the proposed method, extensive parameter studies have been conducted, including the number of image pyramid levels (NP), bi-directional error threshold (BE), search block size (BS), and the maximum number of iterations during tracking (NI). Details of such parameters will be explained in Section 2.6. The results of the parameter studies indicate that, with the recommended parameter range, the RTDT-bolt method can achieve high accuracy (over 90%) in quantifying bolt rotation angle in real time. Moreover, the proposed method can lift the limitation of monitoring rotation greater than 60 degrees. It should be noted that the traditional HT-based method can potentially be used to monitor the total rotation greater than 60 degrees, provided that the HT algorithm can reliably identify the bolt edges, and also the incremental rotation of the bolt between two adjacent frames is less than 60 degrees. However, the processing of the HT algorithm to detect edges of the bolt may not perform well in the relatively complicated situations aforementioned. Even if the HT algorithm can accurately detect edges of the bolt in every frame, the processing of such an algorithm on all the frames is time-consuming, compared to the optical flow tracking algorithms (Nixon, & Aguado, 2019).

## 2 METHODOLOGY

### 2.1 Overview of real-time integrated detection and tracking framework

Figure 1 depicts the integrated method, RTDT-bolt, to robustly monitor the rotation of structural bolts. First, YOLOv3 has been adopted and modified to create a new version of YOLOv3-tiny to localize the bolts with ROI bounding boxes, in the 1st video frame. The Shi-Tomasi algorithm is then employed to identify high-quality FPs within the ROIs for tracking purpose. Second, the optical-flow KLT feature-tracking algorithm is applied to track the FPs generated, from frame to frame. Third, the YOLOv3-tiny algorithm will be integrated with the KLT tracking algorithm to ensure the high performance of tracking. The YOLOv3-tiny detector will generate new ROIs for the bolts, if the number of FPs being tracked falls below a certain threshold (e.g., less than 50% of the initial number of FPs identified). This will not only eliminate the loss of tracking problem due to external environment changes, such as changes of lighting conditions, but also effectively reduce the accumulated error from long-time tracking. Lastly, the total rotation angle of the bolt can be calculated as the sum of the rotation angle determined at each detect-track interval. Specific details of these procedures are discussed in Section 2.2 to Section 2.4.

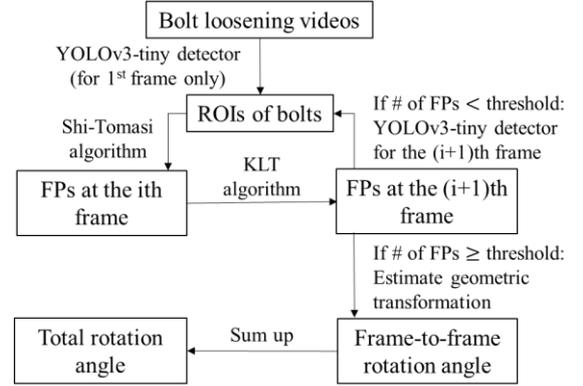

**Figure 1 Flowchart of the RTDT-Bolt method**

### 2.2 YOLOv3-tiny real-time detection networks

#### 2.2.1 Development of YOLOv3-tiny

In this section, the architecture and novelty of the proposed YOLOv3-tiny detector are described. YOLOv3 has been adopted and modified to create a new version of YOLOv3-tiny for localization of bolts. Compared to the original YOLOv3 built on Darknet-53 which has 247 layers in total, the developed YOLOv3-tiny has a total of 44 layers only. This is achieved by reducing the depth in convolutional layers of YOLOv3. The main advantage of YOLOv3-tiny is that it achieves over 10 times higher speed (as will be shown in Section 3.1) than the original YOLOv3, while still maintaining high enough precision for bolt localization. The architecture of the proposed YOLOv3-tiny is depicted in Figure 3. In general, YOLOv3-tiny consists of a series of Convolution, Batch Normalization, Leaky ReLU (Conv-BN-Leaky ReLU) blocks, and max pooling layers. The input image will be resized to 416 x 416 in width and height, before entering the networks for training. During training, YOLOv3-tiny divides the image into 26x26 grid cells. Each grid cell has 3 anchor boxes, of which each has an object score, multiple class scores (depending on the number of classes being detected), and 4 bounding box coordinates. In this case, the number of classes is one, for structural bolts. Consequently, the output of YOLOv3-tiny has a dimension of 26 x 26 x 18, where '26' represents the number of grid cells for each output, and '18' represents each of the three anchor box class scores, object scores, and bounding boxes (i.e., 1 class score for bolt + 4 bounding box values + 1 object score = 6 values. 6 values x 3 anchor boxes = 18 values).

### 2.2.2 Training of RCNN, YOLOv3, and YOLOv3-tiny

The training data were collected in the structural laboratory at the University of British Columbia (UBC). The friction damping device recently developed at UBC was selected to demonstrate the integrated method (Figure 2). First, the bolt rotation was achieved by rotating the bolt on the backside of the damping device. Meanwhile, the videos of bolt rotation were recorded by the iPhone Xs Max smartphone device on the front side, at 4K video quality settings. The phone camera was placed at various angles in order to generate more variety in the dataset. Then, the video frames were processed and extracted to generate the training images for YOLOv3-tiny. Standard data augmentation techniques such as cropping, horizontal flipping, small translations, rotation, and small scaling are also applied such that the object being localized is still included in the augmented images. In this case, 3808 images were generated from the original frames of all video files.

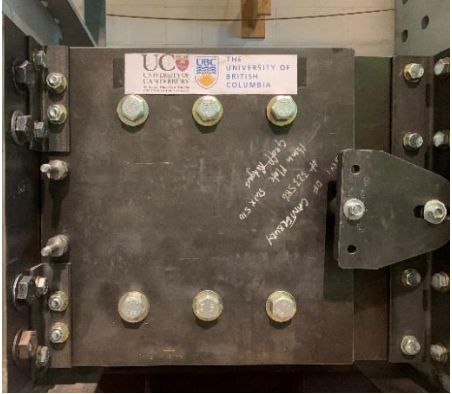

**Figure 2 Image of the experimental friction damper specimen**

The proposed integrated method is designed to deal with illumination changes whose effects have rarely been studied by existing structural engineering vision-based research, but can happen quite often in real-world situations. In order to enhance the data variety and enhance the robustness of the object detector against illumination changes, a lighting-condition-oriented data augmentation approach (Chaichulee et al., 2017) is applied to generate three different lighting conditions to an image. The procedures are briefly explained as follow: a) convert the image from red-green-blue (RGB) space to hue-saturation-lightness (HSL) color space; b) obtain the histogram of average lightness value of all the original images; c) evenly divide the histogram into 3 sections and compute the mean of lightness for each section; d) for each original image falling in one section, two more images are generated by scaling its average lightness to the other two mean of lightness levels in the other two sections, respectively. Details of such augmentation method are described in (Chaichulee et al., 2017). In this regard, the database after the lighting augmentation contains $3808 \times 3 = 11424$ images.

In the end, 70% of the augmented image database is assigned as the training dataset and the rest is selected as the testing dataset. Consequently, $11424 \times 0.7 \approx 7997$, $11424 \times 0.3 \approx 3427$ images are distributed for training, and testing source, respectively. The images are resized to $416 \times 416$, before being input into the networks for training. Then, the selection of anchor boxes is conducted for training data using the methodology presented in Redmon and Farhadi (2018) and Pan and Yang (2020). The anchor box dimensions will be utilized by YOLOv3-tiny to predict the bounding box location for objects in input images. The training of YOLOv3-tiny was implemented by back-propagation and stochastic gradient descent with momentum (SGDM), where the learning rate was chosen as 0.001, the mini-batch size was chosen as 6 and the maximum number of training epochs was set to 80. In addition, to demonstrate the speed and accuracy of the YOLOv3-tiny against the existing object detection algorithms, the RCNN and YOLOv3 detection algorithms are also implemented, respectively. The RCNN algorithm is built on AlexNet, which is adopted from Ta and Kim (2020). Similarly, the training of RCNN was implemented by back-propagation and stochastic gradient descent (SDG), where the learning rate was chosen as 0.000001, the mini-batch size was chosen as 32 and the maximum number of training epochs was set to 10. The YOLOv3 algorithm is built on Darknet-53, which is adopted from Redmon and Farhadi (2018). The training setting of YOLOv3 is the same as that of YOLOv3-tiny. All the training was implemented in MATLAB R2021a (MATLAB, 2021) on two computers: a Lenovo Legion Y740 (a Core i7-8750H @2.20 GHz, 16GB DDR4memory and 8GBmemory GeForce RTX 2070 max-q GPU), an Alienware Aurora R8 (a Core i7-9700K@3.60 GHz, 16 GB DDR4 memory and 8 GB memory GeForce RTX 2070 GPU).

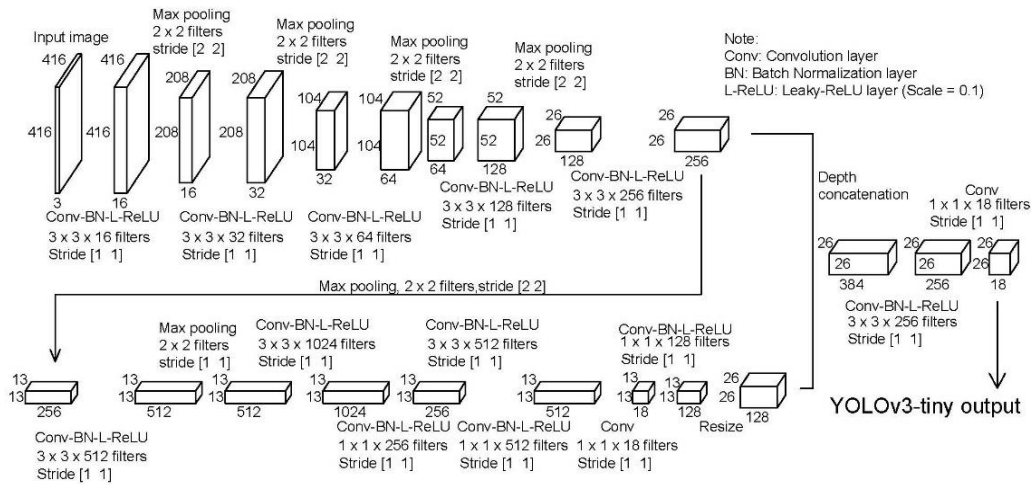

**Figure 3 Architecture of the YOLOv3-tiny object detector**

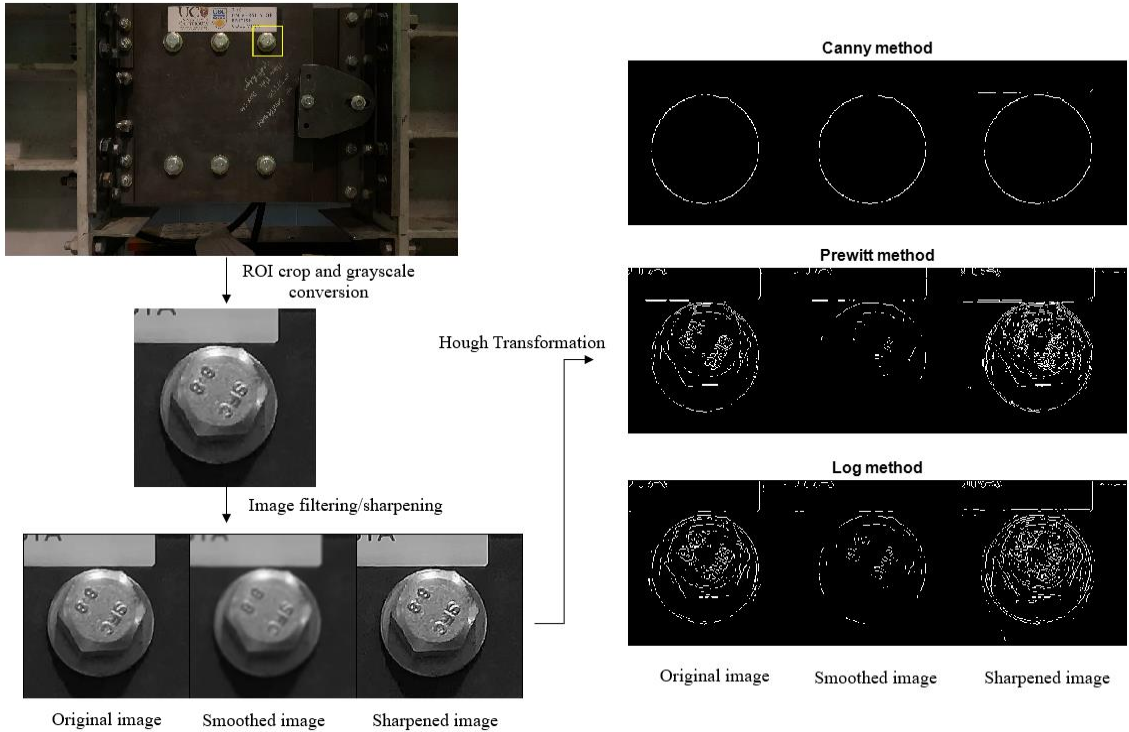

**Figure 4 Hough transformation of the original image, smoothed image, and sharpened image, using Canny method, Prewitt method and Log method, respectively.**

## 2.3 RTDT-Bolt method

In this section, the integrated RTDT-Bolt method, for robust detection and tracking of bolt loosening is described. Earlier research (Park et al., 2015b; Huynh et al., 2019; Ta, & Kim, 2020) has demonstrated the effectiveness of the vision-based methods for bolt rotation estimation. Although their results are promising, these studies have several limitations aforementioned. In particular, the HT algorithm employed in these studies may not be able to accurately detect lines and circles in complex images, e.g., when the image contains washers, light reflections, shades of surrounding objects, or background noise. To illustrate this challenging phenomenon, the HT algorithm using three different methods (i.e., Canny, Prewitt and Log) is applied to three types of images, including the original image, smoothed image, and sharpened image. The original image is cropped from an image of the friction damping device presented. The smoothed image is obtained by applying the Gaussian image filter with a standard deviation of two to the original image, while the sharpened image is generated by subtracting a blurred (unsharp) variant of the image from itself. The original, blurred and sharpened images were processed by the three methods (i.e., Canny, Prewitt and Log), respectively, to identify edges in the images. Then, the HT algorithm is used to identify the straight line edges, which is achieved by collecting the votes of the identified edges by the three methods in the Hough space and selecting the highest votes as the fitted line edges. The procedures were implemented in MATLAB R2021a, where the edge sensitivity threshold for the three HT methods is set as [0.1 0,8], 0.05, and 0.004, respectively. Results in Figure 4 indicate that the identification of the hexagon-shaped edges of the bolts in our experiments is difficult using HT methods. The neighboring object(s), circular-shaped washer, shades, and background noise can cause the algorithm to fail in identifying bolt edges. Besides, although the RCNN method employed by some of these studies provides good localization accuracy, the architecture of the RCNN is relatively heavy. Hence, its speed is too slow for real-time applications when it is desired. On the other hand, the effectiveness of optical-flow-based tracking algorithms has been demonstrated in vision-based structural motion tracking (Ji, & Chang, 2008; Chen et al., 2015; Zheng, Shao, Racic, & Brownjohn, 2016; Cha, Chen, & Büyüköztürk, 2017; Kuddus et al., 2019). Two main limitations have been identified. First, these studies focused on the extraction of horizontal or vertical translations, the investigation of rotation estimation of structural components that exhibit rotation behavior, such as bolts, remains very rare. Second, although these studies have shown promising results, there existed several challenges, such as outdoor lighting conditions. In essence, the optical flow methods are known to work well for tracking objects that have a rigid-body profile and distinct visual texture, but it tends to fail in the situation of sudden external environment changes, such as change of outdoor lighting conditions, light reflection, or shades of neighboring objects (Nixon, & Aguado, 2019). The proposed RTDT-Bolt method is aimed to address these issues. The scope herein is to, a) achieve real-time performance in both detection and tracking; b) provide solutions to measure the rotation of bolts up to any range; c) enhance the robustness of traditional KLT tracking algorithms against illumination changes and background noise.

The detailed implementation of the proposed RTDT-Bolt method is described herein. First, the YOLOv3-tiny algorithm is implemented to generate the bounding box (i.e., ROI) for each bolt in the 1st frame of the video. Second, these ROIs will be extracted from the original video frame and the Shi-Tomasi algorithm is applied to generate FPs inside the ROIs. This step is essential to eliminate the need to process the entire video frame, but rather focus on the ROIs, thus greatly reducing the computational burden. There are two essential parameters involved in the Shi-Tomasi algorithm, including the minimum quality measure, and the Gaussian filter dimension. The minimum quality measure determines the minimum threshold below which the FPs will be discarded. It is recommended to set a reasonably large value to eliminate low-quality points. The Gaussian filter dimension determines the dimension of the Gaussian filter used to smooth the gradient of the input image. In this study, the minimum quality of the Shi-Tomasi FP generation is set as 0.2, while the Gaussian filter dimension is set as 5.

Further, the KLT tracking algorithm initiates on the FPs generated. As the tracking moves forward, some FPs may get lost, due to external environmental changes such as lighting conditions, or the variation of background noise, or the change of the relative location of cameras with respect to the bolts under severe earthquake shaking. In this case, if the number of the FPs being tracked is below a predefined threshold, the tracking is considered lost. In this case, the real-time detector, YOLOv3-tiny, will interfere with the specific frame where the tracking gets lost, and generate new ROIs again. New FPs will be generated inside the ROIs and the tracking continues on the new FPs the same way as before. The above detect-track steps are repeated whenever the tracking is considered lost throughout the videos. Meanwhile, the geometric transformation matrix about the origin, $T$, can be evaluated based on the locations of FPs obtained by the KLT tracking algorithm between two adjacent frames (MathWorks, 2021). Similar to Kuddus et al. (2019), the MSAC algorithm (Torr & Zisserman, 2000) is applied to remove outliers in this step. Then, the rotation angle can be extracted from the transformation matrix using the following steps. Consider a rigid-body object in MATLAB image coordinate system are rotated by $\theta$ angle about the origin, and translated by $t_x$ pixel in horizontal direction, and $t_y$ pixel in vertical direction, which can be expressed by:

$$[x_{i+1}\ y_{i+1}\ 1] = [x_i\ y_i\ 1] * T \quad (1)$$

where $x_{i+1}$, $y_{i+1}$ are the horizontal and vertical pixel coordinates, respectively, of a feature point on the object in the $(i+1)$th frame. Similarly, $x_i$ and $y_i$ are those in the $i$th frame. $T$ is the 2D affine geometric transformation matrix about the origin as defined in MATLAB:

$$T = \begin{bmatrix} cos\theta & sin\theta & 0 \\ -sin\theta & cos\theta & 0 \\ t_x & t_y & 1 \end{bmatrix} \quad (2)$$

Further, the transformation matrix $T^*$ with respect to an arbitrary point (e.g., the center of rotation of bolt) with coordinates of (a, b) can be derived based on $T$,

$$[x_{i+1}\ y_{i+1}\ 1] = [x_i - a\ y_i - b\ 1] * T + (a\ b\ 0) \quad (3)$$

Simplifying equation (3) gives:

$$[x_{i+1}\ y_{i+1}\ 1] = [x_i\ y_i\ 1]$$
$$* \left( \begin{bmatrix} 1 & 0 & 0 \\ 0 & 1 & 0 \\ -a & -b & 1 \end{bmatrix} T + \begin{bmatrix} 0 & 0 & 0 \\ 0 & 0 & 0 \\ a & b & 0 \end{bmatrix} \right) = [x_i\ y_i\ 1] * T^* \quad (4)$$

where,

$$T^* = \begin{bmatrix} 1 & 0 & 0 \\ 0 & 1 & 0 \\ -a & -b & 1 \end{bmatrix} T + \begin{bmatrix} 0 & 0 & 0 \\ 0 & 0 & 0 \\ a & b & 0 \end{bmatrix} \quad (5)$$

It is observed that the 1st row and 2nd row of $T^*$ are the same as those of $T$. Therefore, the incremental bolt rotation angle can be easily extracted from the transformation matrix $T$ estimated. Finally, the incremental rotation estimated at each interval, is summed up to determine the total rotation angle of the bolt, $\varphi$. The time history of the rotation can also be generated. The procedures were implemented in MATLAB R2021a.

### 2.4 Evaluation of the ground truth rotation angle

To assess the feasibility of the proposed integrated method, it is necessary to obtain the ground-truth value for the rotation of the bolt. This can be done using the following simple steps: a) manually label the line edges of the bolts, at an appropriate interval of video frames such that the rotation experienced by the bolt does not exceed 60 degrees within each interval; b) similar to the method presented by Ta and Kim (2020), apply geometric transformation method for all the labeled line edges, and compute the rotation of the bolt, $\theta_{GT,j}$, as the mean rotation of all the line edges, for each interval, $j$; c) sum up the rotation for each interval to determine the total ground-truth rotation angle, $\varphi_{GT}$, as shown below,

$$\varphi_{GT} = \sum_{j=1}^{n=number\ of\ intervals} \theta_{GT,j} \quad (6)$$

In the end, the accuracy of bolt rotation estimation in percentage can be determined as follow (when $\varphi_{GT} \neq 0$),

$$Accuracy = \max\left(0, 1 - \left|\frac{\varphi - \varphi_{GT}}{\varphi_{GT}}\right|\right) \quad (7)$$

where $\varphi$ is the estimated total rotation angle by the RTDT-Bolt method, and $\varphi_{GT}$ is the ground truth rotation angle.

### 2.5 Description of parameter studies

In order to examine the capability and potential limitations of the proposed RTDT-Bolt method, extensive parameter studies are conducted. In this section, the scope of the parameter studies is described. The sensitivity of results to the selection of detection and tracking parameters is reported and discussed in detail.

Given that the KLT algorithm is the essential part of the proposed RTDT-Bolt method, the parameters of the KLT algorithm are investigated with the values shown in Table 1. These parameters are explained as follows:

a) Number of pyramid levels (NP): The KLT tracking algorithm generates an image pyramid, where each subsequent level of the pyramid decreases in resolution by a factor of two compared to the previous level. If the number of pyramid levels is set greater than one, the algorithm tracks the points at multiple levels of resolutions, which may potentially enhance the tracking effectiveness. However, as the computational cost increases with the increase in the number of pyramid levels, it is recommended to select an appropriate value to balance between speed and accuracy.

b) Bi-directional error threshold (BE): The bi-directional error is calculated based on the FPs in the two adjacent frames. Essentially, the algorithm conducts forward-backward tracking. It tracks the FPs from the preceding frame to the current frame, and then traces back the same FPs to the previous frame. The bi-directional error represents the pixel distance in the image coordinate system, between the original location of the points and the backward-tracing location. The FPs will be abandoned when the error associated with them is greater than the threshold.

c) Search block size (BS): This metric determines the neighboring area around the point being tracked. The computational time increases as the block size increases.

d) Maximum number of iterations (NI): This parameter is the maximum number of iterative searches performed by the KLT algorithm to determine the new location of each FP until it converges.

**Table 1 Parameters examined for KLT tracking algorithms**

|    | Examined values | Value of the base model |
|----|-----------------|-------------------------|
| NP | 1,2,3,4         | 3                       |
| BE | 2,6,10,20       | 6                       |
| BS | 5, 11,21,31     | 5                       |
| NI | 10, 20, 30, 40  | 30                      |

The parameter studies are initiated from a base model, whose selection of values for each parameter is also provided in Table 1.

## 3 EXPERIMENTS AND RESULTS

### 3.1 Training and testing results of RCNN, YOLOv3 and YOLOv3-tiny

In this section, the training and testing results of the RCNN, YOLOv3, and YOLOv3-tiny for bolt localization are presented. Six anchor boxes are selected based on the training data and applied for the training of both YOLOv3 and YOLOv3-tiny, using the methodology described in Section 2.2. The dimensions of the anchor boxes are presented in Table 2. Figure 5 provides the precision-recall curves of the three object detectors, for training and testing, respectively. The performance of an object detection algorithm is usually assessed by the precision-recall diagram (Everingham, Van Gool, Williams, Winn, & Zisserman, 2010). In short, a low false positive rate corresponds to a high precision value, and a low false negative rate reflects a high recall value. The overall performance of the algorithm is reflected by the area under the recall-precision curve, where a large value of area indicates the detector has both high recall and precision. In other words, if a detector has high precision but low recall, it can only detect objects in a few sample images, although the localization accuracy is high once the object is recalled. A detector with high recall but low precision can retrieve objects in many images, but the localization error is high inside the images. In the end, the average precision (AP) can be computed from the precision-recall plot, as the weighted average of precision at each recall value. The AP values for both training and testing of the three object detectors are presented in Figure 5. Overall, all the three detectors have achieved high AP during training. However, during testing, YOLOv3 and YOLOv3-tiny show similar performance, while RCNN achieves slightly lower AP. This is because only one class (i.e., bolt) needs to be detected, which does not require an over-complex CNNs such as RCNN and the original YOLOv3 to achieve desired accuracy. Figure 6 provides sample images processed by the YOLOv3-tiny where all the bolts are detected by bounding boxes with high confidence score. In addition, the speed of the three object detectors was also examined by applying the RCNN, YOLOv3 and YOLOv3-tiny, respectively, through the full testing set 5 times. The average speed is calculated as the number of frames or images processed per second (FPS). Table 3 shows a speed comparison of RCNN, YOLOv3 and YOLOv3-tiny, using the Alienware Aurora R8 computer and software platform presented in Section 2.2.2. As shown in Table 3, RCNN method achieves about 0.05 FPS, while YOLOv3 runs at 2.23 FPS. The speed of RCNN is less than the required minimum speed (1.5 FPS, assuming 90 degrees per second of bolt rotational speed), which indicates RCNN is too slow to be adopted in the proposed RTDT method. On the other hand, the speed of YOLOv3 is very close to the minimum required speed. In the situation of slightly lower hardware specs, YOLOv3 cannot achieve real-time speed. The proposed YOLOv3-tiny achieves about 25 FPS, which is about 500 times faster than the RCNN, and about 10 times faster than YOLOv3. This demonstrates the speed and accuracy of YOLOv3-tiny for localizing the steel bolts in real time.

**Table 2 Estimation of anchor box dimensions**

| Index           | 1  | 2  | 3  | 4  | 5  | 6  |
|-----------------|----|----|----|----|----|----|
| Width [pixels]  | 53 | 47 | 36 | 37 | 32 | 29 |
| Height [pixels] | 42 | 37 | 38 | 35 | 33 | 30 |

**Table 3 Speed comparison of RCNN, YOLOv3 and YOLOv3-tiny**

| Object detector | RCNN | YOLOv3 | YOLOv3-tiny |
|-----------------|------|--------|-------------|
| Speed [FPS]     | 0.05 | 2.23   | 25.16       |

### 3.2 The integrated method against illumination changes

As one of the major goals in this study is to deal with the illumination changes, which hampers the application of traditional optical-flow-based tracking algorithms in real-world situations, this section intends to showcase the effectiveness of the proposed RTDT-Bolt method against the illumination changes. The RTDT-Bolt method is implemented with the parameters of the base model. The minimum threshold for the number of FPs is set to 7 in these experiments. If the number of FPs during tracking is below this threshold, the YOLOv3-tiny re-applies detection, and new FPs will be generated for continuous tracking. Besides, to demonstrate the advantages of the integrated method over the traditional optical flow algorithms, the KLT tracking algorithm without the YOLOv3-tiny was also investigated in parallel. Figure 7 (a) illustrates the light-changing scenarios conducted in the laboratory, where the light was switched on and off about every 10 seconds. Figure 7 (b) shows a close-up video frame montage of the bolt being processed by the RTDT-Bolt method. It can be observed the extra light reflection appeared on the surface of the bolt, when the light was switched on, while disappearing if the light was switched off. The experiment results indicate the KLT tracking algorithm without the YOLOv3-tiny instantly lost tracking when the light was switched on for the first time (i.e., at around 350th frame), and all the remaining frames of the bolt cannot be tracked. In comparison, the RTDT-Bolt method can redetect and continue to track the bolt, when the previous tracking got lost due to light change, as shown in Figure 7. The rotation transformation of the points being tracked is imposed on the ROI bounding box for better visualization. The total rotation angle estimated by the RTDT-Bolt method is 12.68 rads in the anti-clockwise direction, which corresponds to the accuracy of 95.1% (with a ground-truth value of 13.25 rads). In addition, the processing speed has also been examined. The proposed method achieves about 17 frames per second on the original 4K video frames, and about 325 frames per second on the cropped video frames (cropped from the 4K video frame using the ROI detected by the YOLOv3-tiny). This demonstrates both the accuracy and real-time speed of the proposed method in monitoring the bolt rotation angle.

## 3.3 Parameter studies

Parameter studies are conducted on the proposed RTDT-Bolt method to assess the sensitivity of the rotation estimation to the selected parameters shown in Table 1. There is a total of 4 x 4 x 4 x 4 = 256 runs. In order to maintain the unique controlled parameter in each set of runs, the light condition is maintained consistently in all the parameter studies. The comparison between estimated rotation of the bolts with the ground truth values in both short and long video scenarios is first summarized in Table 4. The results were obtained by the base model whose parameters are shown in Table 1. The results indicate the proposed model can accurately quantify the rotation of the loosened bolts and non-loosened bolts in both short and long video scenarios. Figure 9 shows a detailed comparison between the time-history rotation of the bolts with the ground truth for bolt 6 in the short video scenario. The corresponding montage of sample video frames is shown in Figure 10, where two reference lines are used to better visualize the tracking process of the bolt rotation. The initial location of the bolt is indicated by the line with smaller line width, while the current rotational position of the bolt is represented by the line with a larger line width. In addition, the rotation angle corresponds to each thumbnail image is indicated at the bottom right of the image, while the frame number is shown at the top left of the image. Figure 11 depicts the effects of these parameters on the rotation estimation results. As shown in Figure 11 (a1), as the number of pyramids increases, the accuracy increases in general, which is particularly well reflected in the situation of medium accuracy due to the large block size (i.e., 31) being used. However, the number of pyramid levels does not have a substantial effect on the rotation estimation, when the accuracy is relatively high already, as depicted in Figure 11 (a2, a3). On the other hand, the effect of the maximum bidirectional error on the estimation accuracy is quite limited in general, as shown in Figure 11 (a2, b, c2, and d2). Figure 11 (c) indicates that the search block size has a great impact on the accuracy of rotation estimation. In general, as the block size increases, the performance of the proposed method decreases. This can be attributed to the fact that the KLT tracking algorithm is more likely to mismatch the points being tracked at the current frame, with the incorrect points in the next frame, when the search area becomes larger where more adjacent (but irrelevant) points are included. Figure 11 (c1) also shows that when the number of pyramid levels is low, as the search block size increases, the accuracy degrades faster than the case where the number of pyramid levels is set higher. However, the search block size will have less impact on the performance, if the maximum bidirectional error, the maximum number of iterations, and the number of pyramid levels are set appropriately, as shown in Figure 11 (c2, c3). In these cases, the accuracy converges to about 90%, with the increase of the block size, maximum bidirectional error, and the maximum number of iterations. Lastly, the maximum number of iterations has minor effects on the results, as shown in Figure 11 (d). This implies that the number of iterations can be set reasonably low to achieve faster speed, if there exist hardware limitations.

In addition, a complete set of results (i.e., 256 runs) obtained from the parameter studies is summarized in Table 5, where the highest and lowest accuracies are highlighted. In these experiments, the run with the highest accuracy (i.e., 99.86%) is achieved, when the number of pyramid levels is set the highest, the maximum bidirectional error and search block size are set the lowest. This is explicable because this set of parameters imposes the most stringent requirements for the algorithm to limit the tracking error. On the other hand, the run with the lowest accuracy (i.e., 46.9%) is observed, where the number of pyramid levels, maximum bidirectional error, search block size, and the maximum number of iterations are set to 1, 20, 31, 40, respectively. This is close to the worst-case scenario among all the experiments, which is also in line with our expectations.

In short, a reasonably high value (i.e., greater than 2) for the number of pyramid levels, and a sufficiently low value (i.e., less than 21) for the search block size must be specified to achieve the desired accuracy (i.e., over 90%). Meanwhile, the maximum number of iterations do not have noticeable effects on the accuracy, and therefore, it can be set at a reasonably low value (e.g., 10) to reduce computational cost. Overall, the parameter studies confirm the effectiveness of the proposed method, and provide recommendations to achieve high accuracy and speed in both short video and long video scenarios. It should be noted that, in the situation of long-term monitoring which can last for months or years, the detection of bolts by the YOLOv3-tiny can be set to apply at an appropriate interval (e.g., every 5 min) to reset the tracking. This will further alleviate the potential accumulated errors due to noise during long-time tracking.

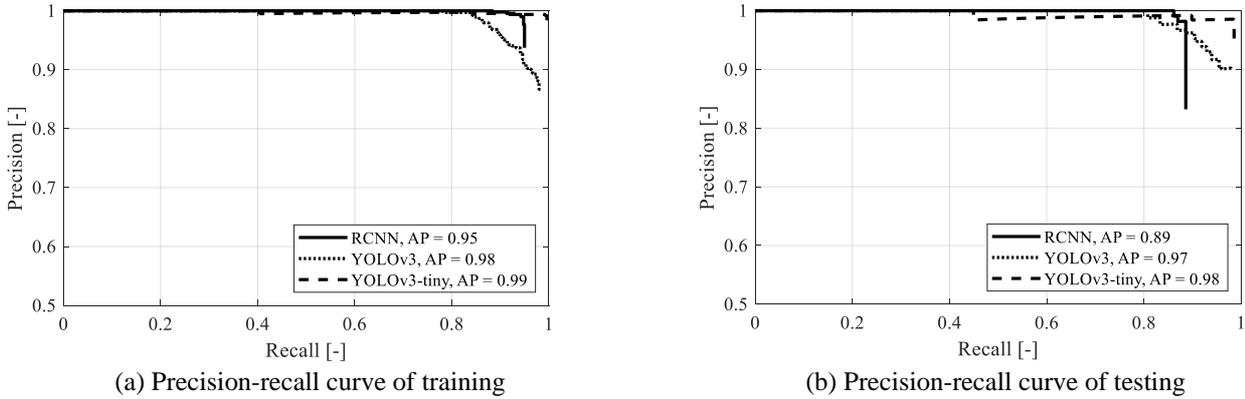

(a) Precision-recall curve of training  (b) Precision-recall curve of testing

**Figure 5 Precision-recall curve for (a) training, and (b) testing**

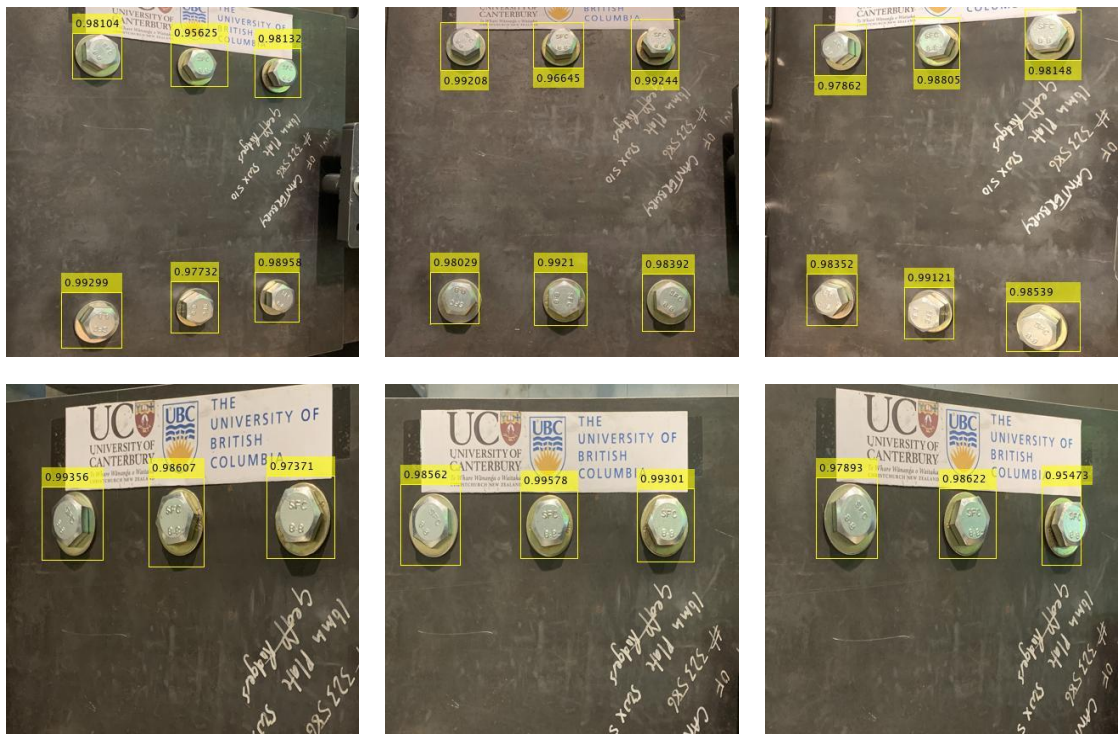

**Figure 6 Sample results of YOLOv3-tiny detection of steel bolts**

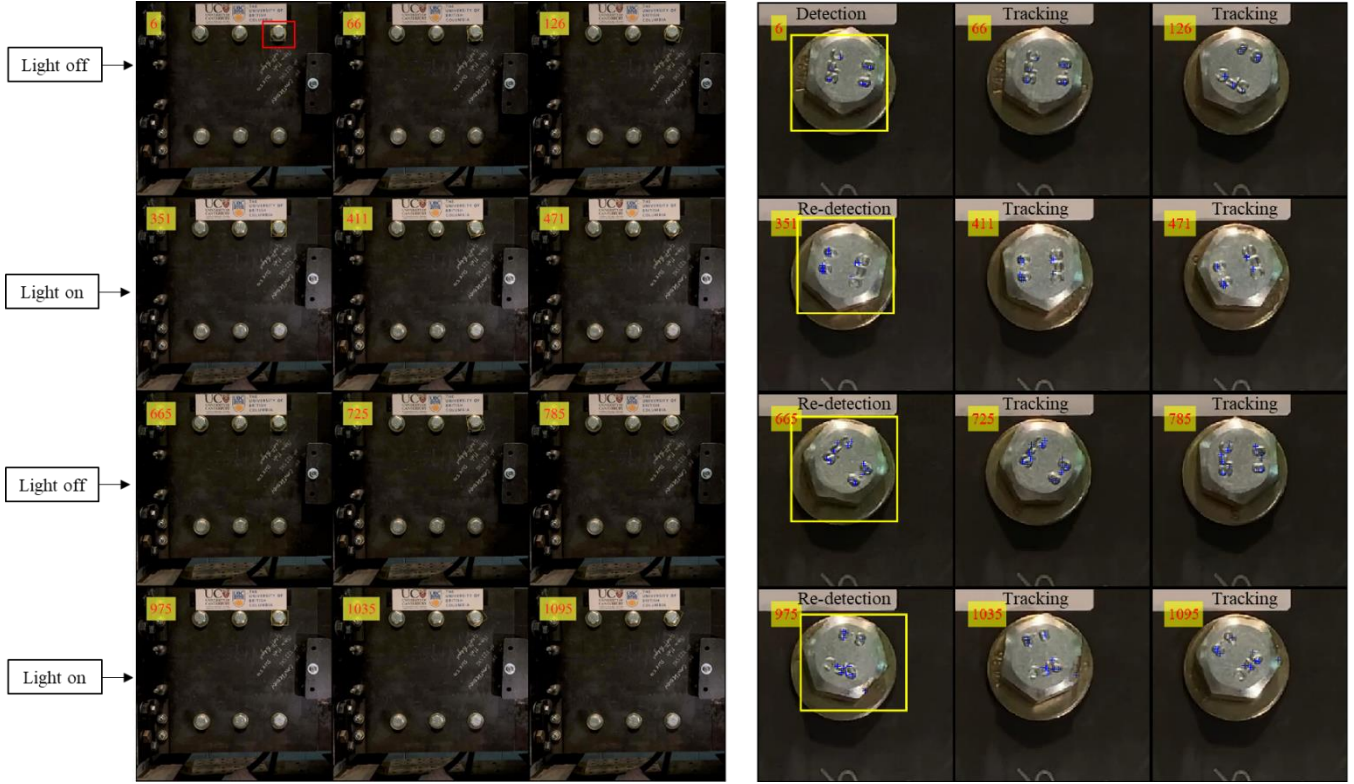

(a)                                                             (b)

**Figure 7 Montage of videos processed by the RTDT-bolt method: (a) original video frame with the illustration of the changing light conditions, and a highlight of the bolt under investigation by the rectangular box; (b) closed-up video frame, with the illustration of detection, tracking, and re-detection. (Note: the frame index is shown at the top-left corner of each thumbnail image. Frame rate: 30 frames per second)**

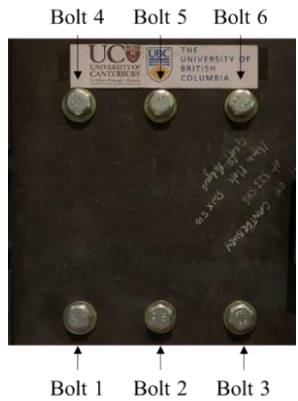

**Figure 8 Illustration of bolt index**

**Table 4 Comparison of the estimated rotation and ground truth rotation for the six bolts in the short video processed by the base model**

|  | Bolt 1 | Bolt 2 | Bolt 3 | Bolt 4 | Bolt 5 | Bolt 6 |
|---|---|---|---|---|---|---|
|  | Short video (441 frames) | | | | | |
| Estimated rotation, [rads] | 0.026 | -0.017 | -0.019 | 0.022 | 0.024 | 8.42 |
| Ground truth rotation, [rads] | 0 | 0 | 0 | 0 | 0 | 8.45 |
|  | Long video (10,504 frames) | | | | | |
| Estimated rotation, [rads] | 0.034 | 0.028 | -0.026 | 0.031 | 0.026 | 51.61 |
| Ground truth rotation, [rads] | 0 | 0 | 0 | 0 | 0 | 54.32 |

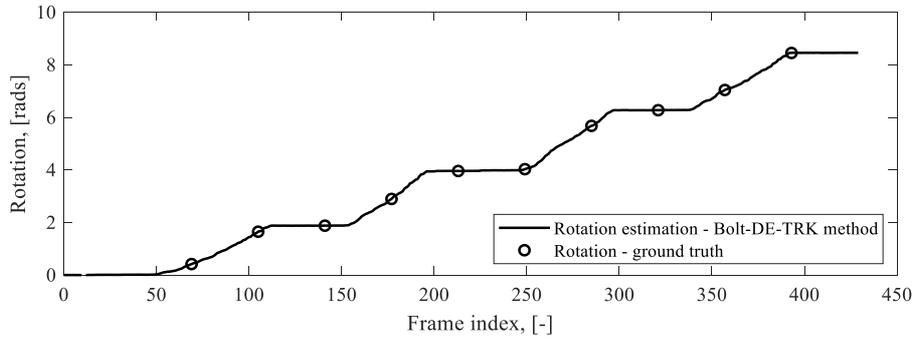

**Figure 9 Time-history rotation estimation of the bolt in the short video with the base model**

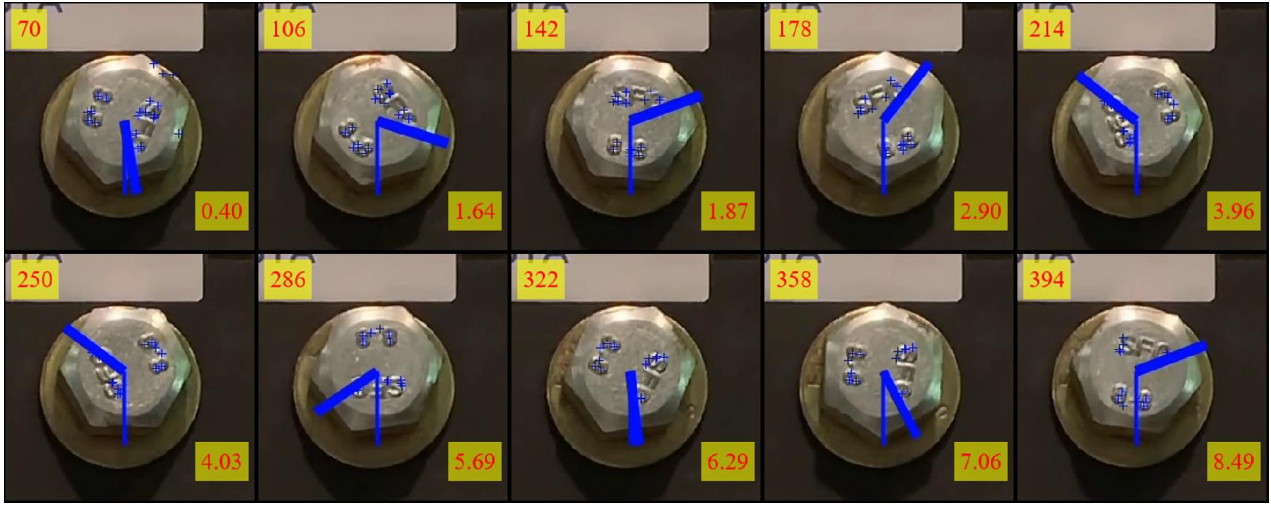

**Figure 10 Montage of sample close-up frames of the short video processed by the base model in parameter studies (Note: each thumbnail image with the labeled frame index corresponds to the associated ground truth point in Figure 9)**

**Table 5 Complete results of parameter studies, expressed by the accuracy of rotation estimation**

| | | NI = 10 | | | | NI = 20 | | | | NI = 30 | | | | NI = 40 | | | |
|---|---|---|---|---|---|---|---|---|---|---|---|---|---|---|---|---|---|
| | | BE = 2 | BE = 6 | BE = 10 | BE = 15 | BE = 2 | BE = 6 | BE = 10 | BE = 15 | BE = 2 | BE = 6 | BE = 10 | BE = 15 | BE = 2 | BE = 6 | BE = 10 | BE = 15 |
| BS = 5 | NP = 1 | 99.81% | 97.81% | 97.83% | 97.19% | 98.34% | 98.31% | 98.83% | 98.78% | 98.63% | 98.95% | 98.89% | 99.06% | 98.56% | 98.92% | 98.83% | 98.74% |
| | NP = 2 | 99.40% | 96.58% | 99.12% | 99.16% | 99.07% | 99.19% | 99.34% | 99.07% | 99.18% | 98.86% | 98.85% | 98.83% | 98.66% | 98.86% | 98.86% | 98.42% |
| | NP = 3 | 99.85% | 98.59% | 97.01% | 97.24% | 99.38% | 99.45% | 99.70% | 99.61% | 99.74% | 99.66% | 99.51% | 99.82% | 99.50% | 99.52% | 99.39% | 99.44% |
| | NP = 4 | 99.49% | 97.79% | 97.78% | 97.68% | 99.85% | 99.43% | 99.58% | 99.42% | 99.86% | 99.32% | 99.54% | 99.64% | 99.62% | 99.41% | 99.48% | 99.59% |
| BS = 11 | NP = 1 | 92.11% | 91.23% | 91.48% | 91.02% | 93.03% | 92.44% | 91.51% | 92.15% | 93.14% | 91.99% | 92.25% | 92.68% | 92.86% | 92.51% | 92.64% | 91.90% |
| | NP = 2 | 92.20% | 91.05% | 91.08% | 90.60% | 92.10% | 93.60% | 91.95% | 92.12% | 92.09% | 92.30% | 92.39% | 92.43% | 92.08% | 92.44% | 92.32% | 92.29% |
| | NP = 3 | 91.81% | 91.13% | 91.39% | 92.89% | 93.38% | 92.12% | 91.80% | 92.44% | 92.01% | 91.97% | 92.41% | 92.38% | 92.42% | 92.31% | 93.03% | 92.40% |
| | NP = 4 | 92.14% | 91.27% | 91.30% | 90.86% | 92.26% | 91.81% | 92.16% | 91.88% | 92.39% | 91.95% | 92.39% | 92.30% | 92.10% | 92.12% | 91.88% | 92.42% |
| BS = 21 | NP = 1 | 88.43% | 87.30% | 88.43% | 87.23% | 92.92% | 87.46% | 87.38% | 87.24% | 88.34% | 87.23% | 87.30% | 92.83% | 92.95% | 89.43% | 88.43% | 87.16% |
| | NP = 2 | 87.78% | 88.40% | 93.61% | 87.45% | 92.79% | 87.82% | 87.27% | 92.85% | 88.22% | 88.35% | 90.49% | 88.01% | 88.56% | 87.32% | 88.34% | 87.12% |
| | NP = 3 | 88.43% | 87.26% | 88.22% | 93.13% | 88.34% | 88.56% | 87.18% | 88.54% | 88.31% | 92.79% | 88.59% | 89.98% | 88.39% | 92.79% | 87.22% | 93.00% |
| | NP = 4 | 88.22% | 92.97% | 87.21% | 88.33% | 87.65% | 88.31% | 87.34% | 92.84% | 84.98% | 88.45% | 92.91% | 88.28% | 88.56% | 88.40% | 92.90% | 88.24% |
| BS = 31 | NP = 1 | 90.58% | 90.58% | 79.84% | 85.54% | 84.67% | 80.27% | 79.63% | 79.91% | 90.58% | 79.57% | 90.58% | 81.62% | 85.07% | 90.58% | 90.58% | 46.90% |
| | NP = 2 | 90.57% | 90.58% | 90.63% | 90.92% | 90.87% | 90.58% | 90.87% | 90.58% | 90.58% | 71.85% | 90.87% | 90.58% | 77.80% | 73.60% | 85.07% | 90.70% |
| | NP = 3 | 85.10% | 90.58% | 90.66% | 90.58% | 84.84% | 90.58% | 90.58% | 90.58% | 90.58% | 90.58% | 85.04% | 90.58% | 90.58% | 90.63% | 73.21% | 85.09% |
| | NP = 4 | 79.77% | 90.57% | 90.57% | 79.55% | 90.57% | 90.57% | 90.57% | 90.57% | 81.79% | 90.57% | 90.63% | 90.86% | 90.56% | 90.57% | 85.33% | 90.57% |

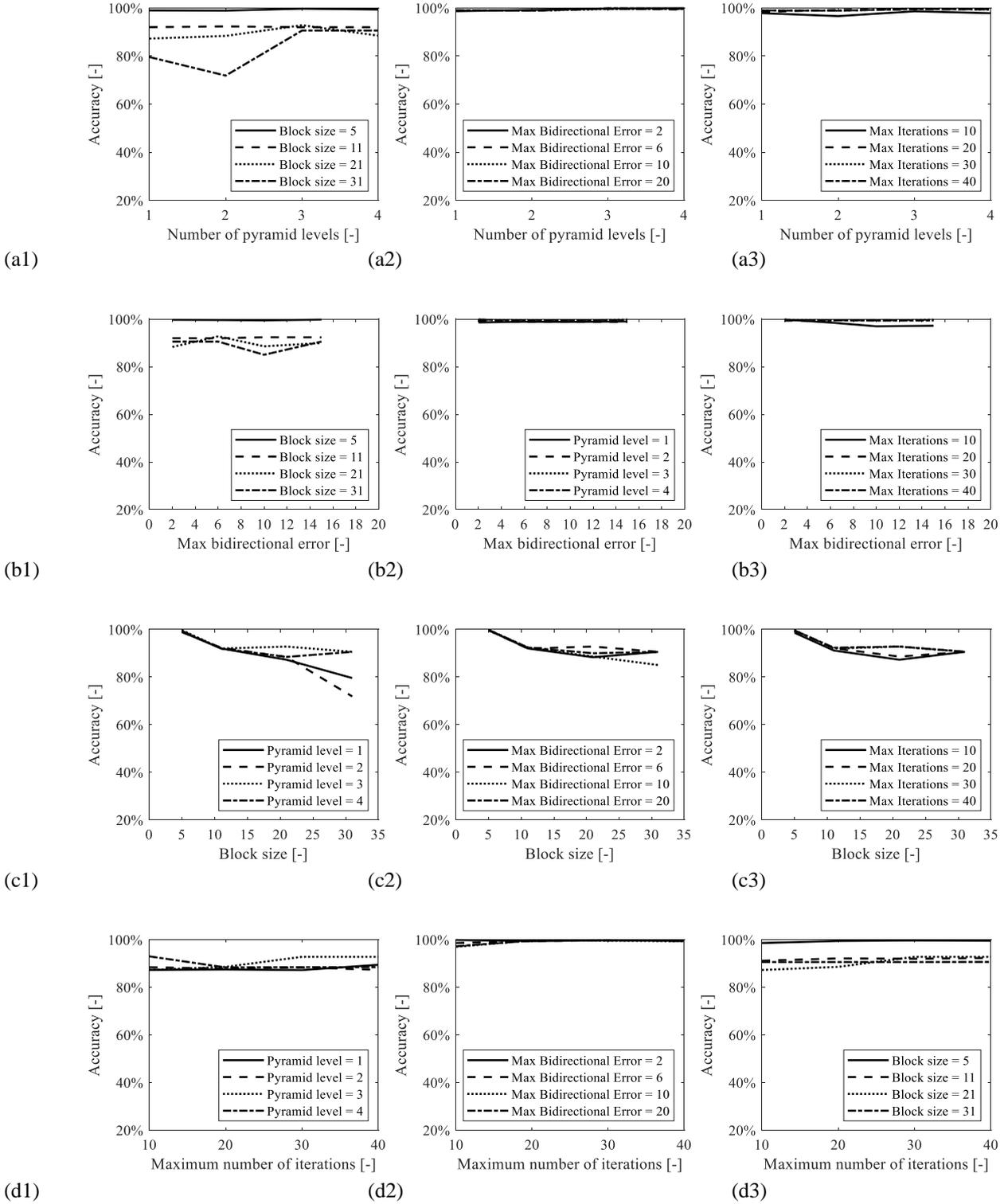

**Figure 11 Sensitivity study of rotation estimation to the selected parameters, with a control parameter of (a1-a3) number of pyramid levels, (b1-b3) maximum bidirectional error, (c1-c3) search block size, and (d1-d3) maximum number of iterations**

## 4 CONCLUSIONS

Structural bolts are commonly used to connect structural components. The forces in the structural bolts are highly dependent on bolt rotation. This article proposed an integrated vision-based method, named RTDT-Bolt, to interactively detect and track the rotation of bolts. The efficient YOLOv3-tiny detector has been established and trained to precisely localize the bolts in real time. Then, the YOLOv3-tiny is integrated with the KLT tracking algorithm to improve the tracking performance. The effectiveness of the proposed method, in dealing with tracking loss problems due to light changes, has been demonstrated over the traditional optical-flow-based tracking algorithms. Further, extensive parameter studies have been conducted to examine the capability and potential limitations of the proposed method. The results indicate the proposed RTDT-Bolt method can reliably quantify the bolt rotation with over 90% accuracy using the recommended range for the parameters. It is also found that the number of pyramid levels and the search block size have great impacts on the rotation estimation, while the maximum number of iterations and the maximum bidirectional error do not have substantial effects on the results. The proposed RTDT method has multiple advantages including: a) achieve real-time performance in both detection and tracking; b) provide solutions to measure the rotation of bolts up to any range; c) enhance the robustness of traditional KLT tracking algorithms against illumination changes and background noise.

# ACKNOWLEDGMENT

The authors would like to acknowledge the funding provided by the International Joint Research Laboratory of Earthquake Engineering (ILEE), National Natural Science Foundation of China (grant number: 51778486), Natural Sciences and Engineering Research Council (NSERC), China Scholarship Council. Any opinions, findings, and conclusions, or recommendations expressed in this paper are those of the authors.